\documentclass{article}

\usepackage{PRIMEarxiv}

\usepackage[utf8]{inputenc} % allow utf-8 input
\usepackage[T1]{fontenc}    % use 8-bit T1 fonts
\usepackage{hyperref}       % hyperlinks
\usepackage{url}            % simple URL typesetting
\usepackage{booktabs}       % professional-quality tables
\usepackage{amsfonts}       % blackboard math symbols
\usepackage{nicefrac}       % compact symbols for 1/2, etc.
\usepackage{microtype}      % microtypography
\usepackage{lipsum}
\usepackage{fancyhdr}       % header
\usepackage{graphicx}       % graphics
\graphicspath{{media/}}     % organize your images and other figures under media/ folder

\usepackage{multirow}
\usepackage{amsmath}    

\usepackage{cleveref}

\newcommand{\vect}[1]{\mathbf{#1}}

\newcommand{\reals}{\mathbb{R}}
\newcommand{\Ltotal}{\mathcal{L}_{\text{total}}}

\newcommand{\Lseg}{\mathcal{L}_{\text{Seg}}}

\def\BibTeX{{\rm B\kern-.05em{\sc i\kern-.025em b}\kern-.08em
    T\kern-.1667em\lower.7ex\hbox{E}\kern-.125emX}}

\begin{document}
\title{TARDis: Time Attenuated Representation Disentanglement for Incomplete Multi-Modal Tumor Segmentation and Classification}
\author{Zishuo Wan \footnotemark[1], Qinqin Kang \footnotemark[1], Na Li, Yi Huang, Qianru Zhang, Le Lu, Yun Bian \footnotemark[2], Dawei Ding \footnotemark[2], and Ke Yan \footnotemark[2]}
\footnotetext[1]{\textit{Zishuo Wan and Qinqin Kang are co-first authors.}}
\footnotetext[2]{\textit{Corresponding authors: Yun Bian, Dawei Ding, Ke Yan.}}
\footnotetext[3]{Zishuo Wan, and Da-Wei Ding are with the School of Automation and Electrical Engineering, University of Science and Technology Beijing, Beijing 100083, China (e-mail: d202410372@xs.ustb.edu.cn; dingdawei@ustb.edu.cn). }
\footnotetext[4]{Yi Huang, and Ke Yan are with Alibaba Group DAMO Academy, Beijing 100026, China, and also with Hupan Lab, Hangzhou 310023, China (e-mail: hy456523@alibaba-inc.com, yanke.yan@alibaba-inc.com).}
\footnotetext[5]{Le Lu is with Alibaba Group DAMO Academy, Beijing 100026, China. }
\footnotetext[6]{Qinqin Kang, Na Li, Qianru Zhang, and Yun Bian are with the Departments of Radiology, Changhai Hospital, Shanghai, China (e-mail: bianyun2012@foxmail.com). }

\maketitle
\begin{abstract}
The accurate diagnosis and segmentation of tumors in contrast-enhanced Computed Tomography (CT) are fundamentally driven by the distinctive hemodynamic profiles of contrast agents over time. However, in real-world clinical practice, complete temporal dynamics are often hard to capture by strict radiation dose limits and inconsistent acquisition protocols across institutions, leading to a prevalent missing modality problem. Existing deep learning approaches typically treat missing phases as absent independent channels, ignoring the inherent temporal continuity of hemodynamics. In this work, we propose Time Attenuated Representation Disentanglement (TARDis), a novel physics-aware framework that redefines missing modalities as missing sample points on a continuous Time-Attenuation Curve. We first hypothesize that the latent feature can be disentangled into a time-invariant static component (anatomy) and a time-dependent dynamic component (perfusion). We achieve this via a dual-path architecture: a quantization-based path using a learnable embedding dictionary to extract consistent anatomical structures, and a probabilistic path using a Hemodynamic Conditional Variational Autoencoder to model dynamic enhancement conditioned on the estimated scan time. This design allows the network to infer missing hemodynamic features by sampling from the learned latent distribution. Extensive experiments on a large-scale multi-modal private abdominal CT dataset (2,282 patients) and two public datasets demonstrate that TARDis significantly outperforms state-of-the-art incomplete modality frameworks. Notably, our method maintains robust diagnostic performance even in extreme data-sparsity scenarios, highlighting its potential for reducing radiation exposure while maintaining diagnostic precision.
\end{abstract}


\section{Introduction}
\label{sec:intro}
Tumor segmentation in contrast-enhanced Computed Tomography (CT) is fundamentally dependent on the transient physiological dynamics of the contrast agent \cite{bi2019artificial}. Crucially, the diagnostic boundaries of tumors are often defined by this time-dependent contrast-enhancement signal, which is unavailable in non-contrast acquisitions or incorrectly timed scans \cite{bae2010intravenous}. Accurately capturing these dynamics usually requires a multi-phase acquisition protocol. 

However, obtaining a standardized, complete set of phases is rarely feasible in real-world clinical practice. First, strict limits on radiation dose and scanning time often force clinicians to skip specific phases, creating a significant missing modality problem. Second, and perhaps more challenging, is the variability of acquisition protocols across different medical centers and tumor types. The definition of a specific phase often relies on fixed time delays that do not account for patient-specific cardiac output or variations in scanner hardware. As a result, a scan labeled as \textit{arterial} at one hospital may effectively be a \textit{late arterial} or \textit{early venous} phase at another. This temporal mismatch introduces severe domain shifts, where the intensity distribution of the same modality varies drastically between datasets. Standard deep learning frameworks, which assume fixed, synchronized input channels, degrade severely when facing such inconsistent or incomplete data \cite{javed2024robustness}.

Existing approaches to multi-modal segmentation typically rely on channel-concatenation strategies, which assume a fixed set of inputs. When a modality is missing, these models often fail or require retraining. To address this, recent incomplete modality frameworks have been proposed, utilizing techniques such as arithmetic mean fusion, knowledge distillation, or shared latent space mapping \cite{dou2020unpaired,chen2019robust}. While these methods allow for flexible inputs, they generally treat different contrast phases as independent, uncorrelated channels or simple statistical variations. By ignoring the temporal dependency between phases, existing methods struggle to infer the missing dynamic information effectively, often resulting in suboptimal feature representation \cite{shen2019brain,sharma2019missing}.

\begin{figure}[tb]
  \centering
  % \fbox{\rule{0pt}{2in} \rule{0.95\textwidth}{0pt}}
  % Make sure the path 'figs/TAC.jpg' is correct in your project
  \includegraphics[width=0.65\columnwidth]{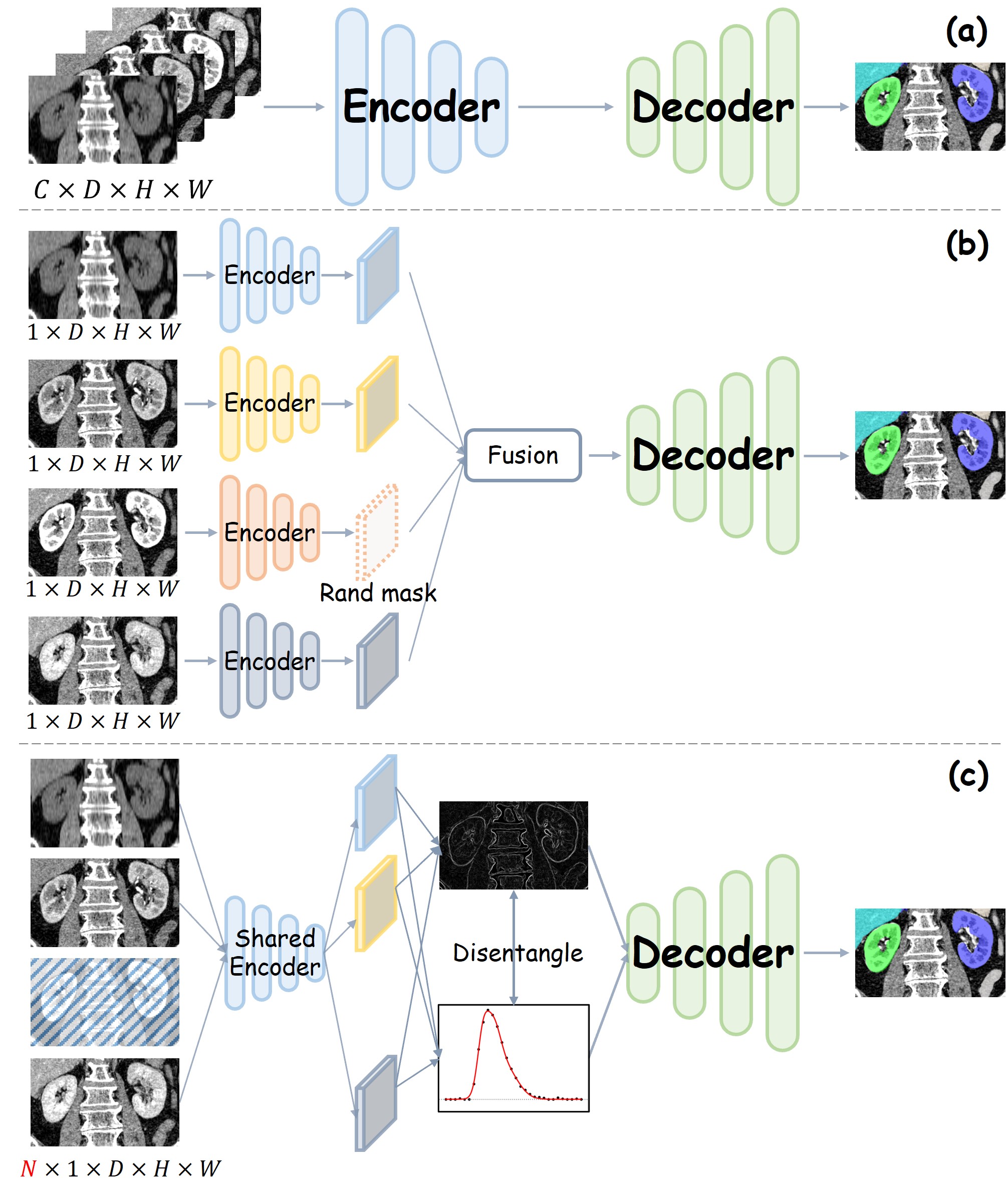}
    \caption{Comparison of segmentation frameworks. (a) Standard U-Net takes a fixed number of modalities. (b) Existing incomplete modality methods that use separate encoders for fixed modalities and fuse them by randomly masking. (c) Our proposed method TARDis, which takes a flexible number of modalities ($N$), processes them via a shared encoder, and explicitly disentangles them into modal-agnostic (anatomic) and modal-specific (temporal) representations before decoding.}
  
  \label{fig:comp}
\end{figure}

In this work, we propose Time Attenuated Representation Disentanglement (TARDis), a novel framework that addresses the incomplete modality challenge by modeling the underlying hemodynamics of CT. Our central hypothesis is that the Hounsfield Unit (HU) intensity can be explicitly decoupled into a time-invariant static component and a time-dependent dynamic component. Rather than treating missing modalities as empty data channels, we treat them as missing samples on the Time-Attenuation Curve (TAC). Our method employs a shared encoder to process a flexible number of input modalities. To learn robust representations, we introduce a dual-path disentanglement mechanism, as shown in Fig.~\ref{fig:comp}. First, we extract a modal-agnostic anatomical representation using a learnable embedding dictionary, ensuring consistency across different scans. Second, we model the modal-specific dynamic representation by designing a physics-aware Conditional Variational Autoencoder (CVAE) \cite{sohn2015learning}, which is conditioned on the anatomy and the estimated scan time. This allows the network to probabilistically reconstruct the dynamic enhancement features of the tumor even when specific phases are missing. By explicitly disentangling anatomy from time-dependent perfusion, TARDis learns a physically grounded latent space that is robust to missing data, superior to standard fusion or masking strategies.

\begin{figure}[tb]
  \centering
  % \fbox{\rule{0pt}{2in} \rule{0.95\textwidth}{0pt}}
  % Make sure the path 'figs/TAC.jpg' is correct in your project
  \includegraphics[width=0.4\columnwidth]{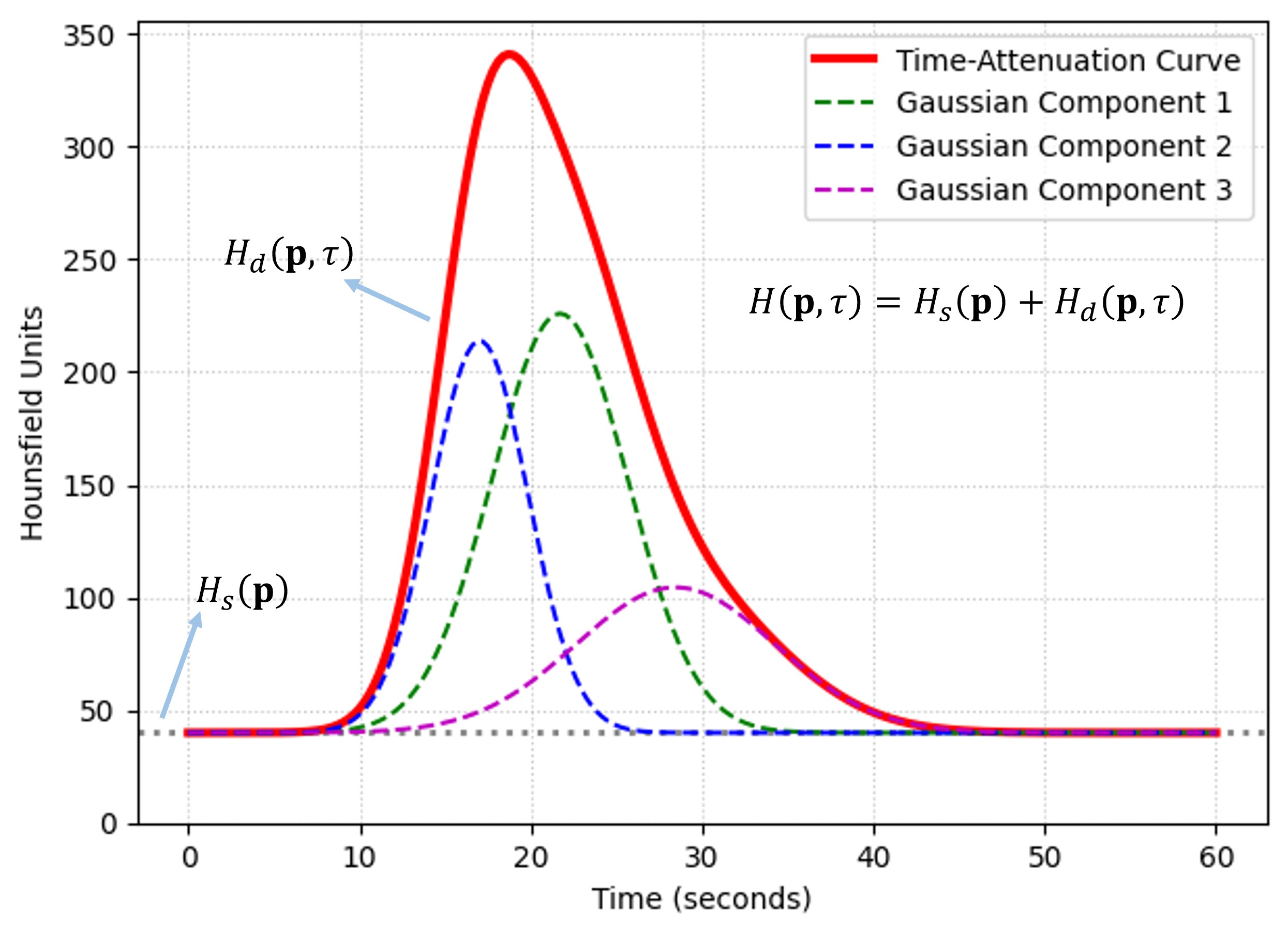}
  \caption{The Time-Attenuation Curve of a certain spatial location $\vect{p}$ in a CT volume. The static component $H_s(\vect{p})$ is related to the tissue's natural density. And the dynamic component $H_d(\vect{p}, \tau)$ reflects how the Hounsfield Unit changes as the contrast agent passes the tissue, which can be approximated with several Gaussian distributions.}
  \label{fig:tac}
\end{figure}

\begin{figure*}[tb]
  \centering
  % \fbox{\rule{0pt}{3in} \rule{0.95\columnwidth}{0pt}}
  \includegraphics[width=0.95\textwidth]{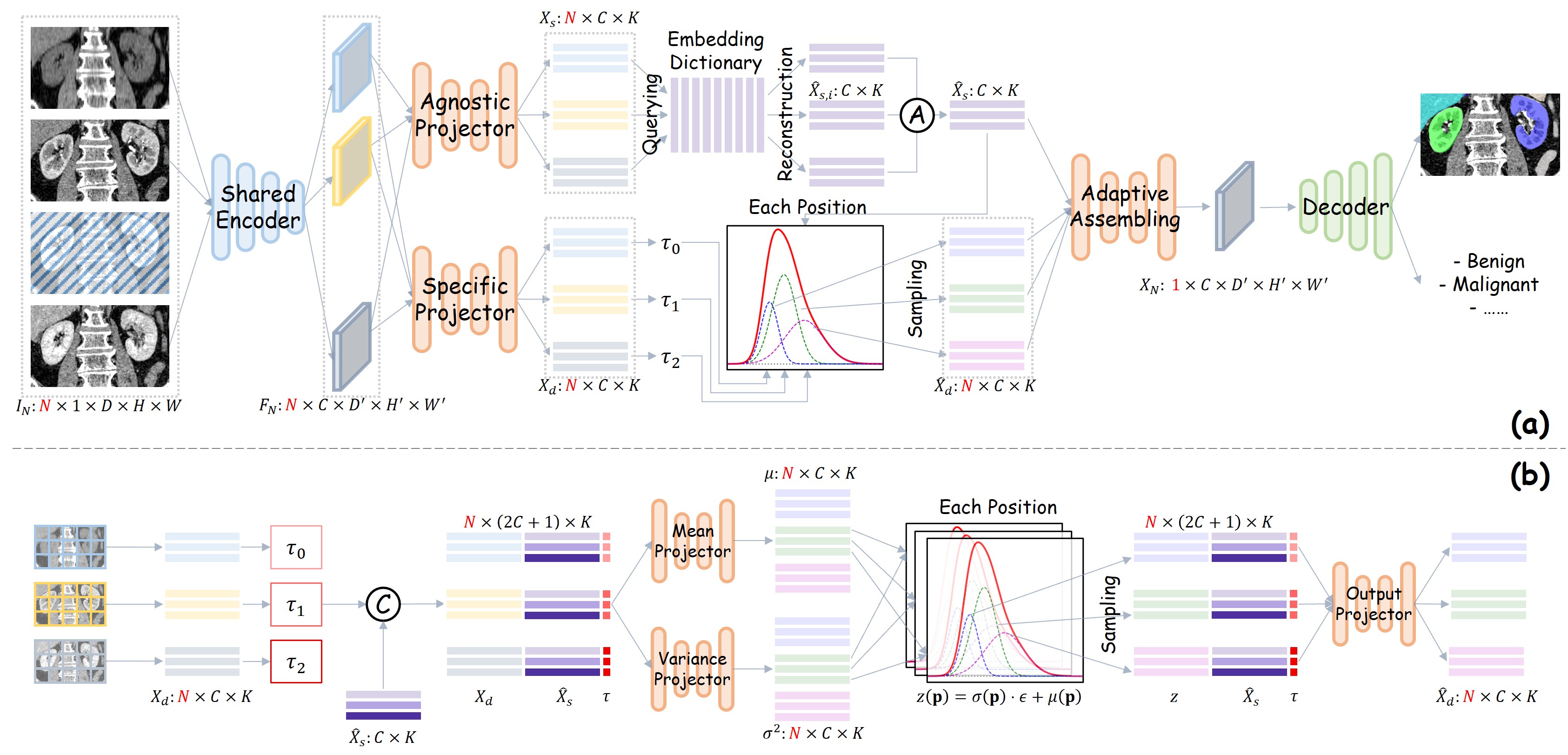}
  \caption{(a) The overall architecture of TARDis. Input volumes are processed by a shared encoder. The features are projected and flattened into $K$ tokens. The modal-agnostic path (top) queries an embedding dictionary to extract consistent anatomic features. The modal-specific path (bottom) regresses a relative time $\tau$ and uses a Hemodynamic Conditional Variational Autoencoder (HCVAE) conditioned on the average anatomy and $\tau$ to sample dynamic features. Finally, the aggregated representations are assembled and decoded by a U-Net decoder. (b) The detailed workflow of the modal-specific path. Every token $X_{d,i}(\vect{p})$ contains the dynamic features of a neighboring subregion in different modalities. First, a $\tau_i$ is regressed based on dynamic features. Then we treat the time $\tau_i$ and modal-agnostic anatomic features $\hat{X}_s(\vect{p})$ as the conditions, and project the dynamic features of each position $\vect{p}$ into Gaussian means $\mu (\vect{p})$ and variances $\sigma^2 (\vect{p})$. Then we reparameterize the tokens with $\epsilon(\vect{p}) \in \mathcal{N}(0, I)$ and reconstruct the $\hat{X}_{d,i}(\vect{p})$. When inferring, we can sample $z(\vect{p})$ directly from $\mathcal{N}(0, I)$ and generate the corresponding dynamic representations with respect to the given $(\hat{X}_s(\vect{p}),\tau_i)$}
  
  \label{fig:architecture}
\end{figure*}

\section{Related Works}
\label{sec:works}
\subsection{Multi-Modal Medical Image Segmentation}
Multi-modal learning has become a cornerstone in medical image analysis, as combining complementary information from different modalities significantly improves diagnostic precision \cite{9363915}. Early approaches focused on input-level fusion, where registered modalities are concatenated as multi-channel inputs to a Convolutional Neural Network \cite{kamnitsas2017efficient,milletari2016v}. While effective, this assumes a fixed set of inputs, making the model fragile to missing data. Feature-level fusion strategies have subsequently emerged to better exploit cross-modal correlations. For instance, Dolz et al. \cite{dolz2018hyperdense} proposed HyperDenseNet to facilitate dense connectivity across multi-modal streams. More recently, Transformer-based architectures have been adopted to model long-range dependencies across modalities. Hatamizadeh et al. \cite{hatamizadeh2022unetr} introduced UNETR, which utilizes a vision transformer encoder to fuse volumetric data. Despite their success, these complete-modality frameworks rely on the assumption that all training and testing modalities are available. In clinical practice—particularly in multi-phase CT—scans are often incomplete due to radiation dose concerns or varying acquisition protocols, rendering these rigid fusion models ineffective without retraining or zero-filling strategies that often introduce artifacts.

\subsection{Learning with Incomplete Modalities}
To address the missing modality problem, recent research has shifted toward flexible frameworks capable of handling arbitrary subsets of inputs. The pioneering work of HeMIS \cite{havaei2016hemis} proposed learning a shared latent representation where first- and second-order statistics of available modalities are fused. This statistical fusion concept was extended by Dorent et al. \cite{dorent2019hetero} in the U-HVED framework, which integrates a multi-modal VAE to reconstruct missing inputs. Other approaches employ knowledge distillation (KD) to transfer information from a complete-modality teacher to an incomplete-modality student \cite{hu2020knowledge,wang2021acn}. However, KD methods typically require a complete set of modalities during the training phase, which may not always be available in retrospective cohorts.
More recently, network architecture designs have moved toward adaptive feature fusion. RFNet \cite{ding2021rfnet} utilizes region-aware fusion to selectively aggregate features based on image content. With the rise of Transformers, methods like mmFormer \cite{zhang2022mmformer} and M2FTrans \cite{m2ftrans} employ masked attention mechanisms to learn intra- and inter-modal correlations, effectively masking out missing tokens. M3AE \cite{liu2023m3ae} adapts the Masked Autoencoder (MAE) paradigm to multi-modal settings. While these methods achieve state-of-the-art results, they generally treat modalities as independent, uncorrelated channels. 

\subsection{Representation Disentanglement and Generative Modeling}
Representation disentanglement aims to decompose data into distinct, explanatory factors, typically separating spatial structure from appearance. In natural image synthesis, methods like MUNIT \cite{huang2018multimodal} and DRIT++ \cite{lee2018diverse} successfully disentangle content and style codes. In medical imaging, this paradigm has been applied to improve robustness and interpretability. Chartsias et al. \cite{chartsias2017multimodal} utilized disentanglement for synthesizing missing MRI sequences. SDNet \cite{chartsias2019disentangled} further decomposes medical images into spatial anatomy and non-spatial modality factors to improve segmentation generalization. VQ-VAE \cite{van2017neural} demonstrates that quantizing latent features can effectively remove noise and preserve essential structural information. In the context of missing data, Generative Adversarial Networks and VAEs are often used to impute missing modalities before segmentation \cite{zhou2020hi,nie2017medical}. However, most existing generative imputation methods perform image-to-image translation without explicit physical grounding.

\section{Method}
\label{sec:method}

\subsection{Problem Formulation}
\label{sec:problem_formulation}

To address the challenges of incomplete modalities, we propose a method based on a decomposition of the Hounsfield Unit (HU) value. In multi-phase CT, the HU value at any spatial location $\vect{p} = (z, x, y)$ is a function of the time $\tau$ elapsed since the injection of a contrast agent. This function, $H(\vect{p}, \tau)$, is known as the Time-Attenuation Curve and describes the complete perfusion profile of the tissue at that location, as shown in Fig.~\ref{fig:tac}. Clinically, a \textit{modality} (e.g., Arterial, Venous, Delayed phase) corresponds to a single sample of this curve at a specific time $\tau$. A non-contrast CT represents the baseline at $\tau=0$.

Our central hypothesis is that the TAC can be explicitly disentangled into two components: a static component $H_s(\vect{p})$ and a dynamic component $H_d(\vect{p}, \tau)$, which can be formulated as:
\begin{equation}
H(\vect{p}, \tau) = H_s(\vect{p}) + H_d(\vect{p}, \tau).
\label{eq:hu_decomp}
\end{equation}
Here, $H_s(\vect{p})$ is the time-invariant static component, which reflects the intrinsic anatomical properties and base density of the tissue at location $\vect{p}$. It is equivalent to the non-contrast value, $H_s(\vect{p}) = H(\vect{p}, 0)$. Since this anatomical information is shared across all phases, we treat it as the modal-agnostic representation. Conversely, $H_d(\vect{p}, \tau)$ is the time-dependent dynamic component. It captures the enhancement profile as the contrast agent perfuses the tissue. This component is, by definition, modal-specific, as it dictates the unique appearance of each contrast phase.

Based on this physical decomposition, our computational goal is to learn a disentangled latent representation for a given position $\vect{p}$ that separates its static properties from its dynamic properties. We propose a model that learns to map multi-modal input CT volumes $I_N$ to a pair of latent representations:
\begin{enumerate}
    \item A modal-agnostic representation $X_s$, which is discrete and aims to capture the intrinsic, time-invariant anatomical properties analogous to $H_s(\vect{p})$.
    \item A modal-specific representation $X_d$, which is continuous and probabilistic, aiming to capture the time-dependent enhancement profile analogous to $H_d(\vect{p}, \tau)$.
\end{enumerate}

We further hypothesize that the dynamic enhancement profile $H_d$ is conditional on the underlying anatomical properties $H_s$:
\begin{equation}
    H_d(\vect{p}, \tau) \sim q(H_s(\vect{p}),\tau).
\label{eq:hypo}
\end{equation}
Therefore, TARDis is designed such that the generation of the modal-specific representation $X_d$ is explicitly conditioned on the modal-agnostic representation $X_s$ and the phase time $\tau$. The final, disentangled representation $X = \{X_s, X_d\}$ is then used for the downstream segmentation task. The details will be given in the following sections.

\subsection{Overall Architecture}
\label{sec:architecture}

The overall architecture of TARDis is illustrated in Fig.~\ref{fig:architecture}(a). Unlike previous incomplete modality methods that use separate encoders for fixed inputs, we employ a shared encoder to extract features from a flexible number of modalities. When taking a series of registered incomplete multi-modal CT volumes $I_N$ as inputs, each of the input modality volumes $I_i \in \reals^{1 \times D \times H \times W}, i \in \{1, 2, \dots, N\}$ is independently processed by a shared 3D encoder, producing a feature map $F_i \in \reals^{C \times D' \times H' \times W'}$. These multi-modal features are then disentangled by a pair of projectors and flattened into sequences of tokens $X_{s} \in \reals^{N \times C \times K}$ and $X_{d} \in \reals^{N \times C \times K}$. Then the tokens are processed by two distinct paths. The modal-agnostic path extracts a common static anatomical representation $\hat{X}_{s}\in \reals^{C \times K}$ using a shared embedding dictionary (Sec.~\ref{sec:modal_agnostic}). The modal-specific path estimates the phase time $\tau_i$ and uses a Hemodynamic CVAE (HCVAE) to model and sample the dynamic representations $\hat{X}_{d} \in \reals^{N \times C \times K}$ for all modalities (Sec.~\ref{sec:modal_specific}). The derived anatomic representation and the multi-modal dynamic representations are adaptively assembled (Sec.~\ref{sec:assembly}), reshaped back to spatial dimensions, and passed to a decoder to predict the final segmentation mask. The entire network is trained jointly using the objectives detailed in Sec.~\ref{sec:loss}.

\subsection{Modal-Agnostic Representations}
\label{sec:modal_agnostic}

The static component represents the inherent anatomy, which should be consistent across all phases. However, features extracted from different contrast phases often contain phase-specific noise that obscures this underlying anatomy. We employ VQ-VAE \cite{van2017neural} for the anatomical branch because anatomy is deterministic and time-invariant. By forcing features to map to a finite, discrete Embedding Dictionary, the VQ mechanism acts as a robust information bottleneck. It effectively filters out phase-specific intensity variations and noise, ensuring that the extracted anatomical representation remains consistent regardless of whether the input is an arterial, venous, or non-contrast scan. We construct the learnable Embedding Dictionary $\mathcal{D} \in \reals^{M \times C}$, containing $M$ anatomical vision queries, into which the anatomical information is distilled by quantization.

For each input modality $i$, the projected static feature at any position $X_{s,i}(\vect{p}_k) \in \reals^{C \times K}$ is used to find the nearest query in the dictionary, and reconstructed by the discrete queries:
\begin{equation}
\hat{X}_{s,i}(\vect{p}_k)=q_k, \quad  k=\underset{q_j \in \mathcal{D}}{\operatorname{argmin}}\left\|X_{s,i}(\vect{p}_k)-q_j\right\|_2
\end{equation}
 where $k\in \{0,1,\dots,K-1\}$. This process reconstructs a refined multi-modal feature $\hat{X}_{s,N}$ by aggregating relevant queries from $\mathcal{D}$. To constrain the consistency of the anatomical information in multi-modal representations, we introduce pair-wise mean squared loss to the querying results:
\begin{equation}
    \mathcal{L_\text{Consis}} = \sum_{i \neq j} (\hat{X}_{s,i}(\vect{p}) - \hat{X}_{s,j}(\vect{p}))^2.
\label{eq:lconsis}
\end{equation}
% This query mechanism acts as a denoising bottleneck, filtering out modal-specific variations and retaining only the information that matches the learned anatomical priors in the dictionary.
% \begin{equation}
% \hat{\vect{X}}_{s,i} = \text{Query}(\vect{X}_{s,i}, \mathcal{D})
% \end{equation}
This constrains the individual embeddings $\hat{\vect{X}}_{s,i}$ from different phases to be close to this global average (and thus to each other), regardless of the input modality. This explicitly aligns the latent spaces of different phases into a unified anatomical space. By constraining, the reconstructed features $\hat{X}_{s,i}$ should be identical for all $i \in \{1,2,\dots,N\}$. To enforce this, we compute the mean of these reconstructed features to obtain a robust, global anatomic representation:
\begin{equation}
\hat{X}_{s}(\vect{p}) = \frac{1}{N} \sum_{i=1}^{N} \hat{\vect{X}}_{s,i}(\vect{p}),
\end{equation}
which is also considered as the conditional information to construct dynamic information. The overall loss of the modal-agnostic path can be written as:
\begin{equation}
    \mathcal{L_\text{Agn}} = \mathcal{L_\text{Consis}} + \| \operatorname{sg}[\hat{X}_{s,i}] - X_{s,i} \|_2^2 + \beta \| \hat{X}_{s,i} - \operatorname{sg}[X_{s,i}] \|_2^2,
\label{eq:lagn}
\end{equation}
where $\operatorname{sg}[\cdot]$ means stop gradient.
% During optimization, we apply a consistency loss $\mathcal{L}_{\text{consist}}$ using Mean Squared Error (MSE). This constrains the individual embeddings $\hat{\vect{X}}_{s,i}$ from different phases to be close to this global average (and thus to each other), regardless of the input modality. This explicitly aligns the latent spaces of different phases into a unified anatomical space.

\subsection{Sampling Based Modal-Specific Representations}
\label{sec:modal_specific}
The dynamic component varies significantly with time, reflecting the perfusion of contrast agents. To model this complex, time-dependent distribution, we design a generative module based on an HCVAE. The details are shown in Fig.~\ref{fig:architecture}(b). This allows us to sample specific representations conditioned on the underlying anatomy and the particular time point.

We first must estimate where the current scan lies on the TAC curve, as the exact acquisition time is often unknown. We regress a relative time scalar $\tau_i$ directly from the modal-specific feature $X_{d,i}$ using a lightweight regressor network $\mathcal{R}$:
\begin{equation}
\tau_i = \mathcal{R}(X_{d,i}) \in [0, 1],
\end{equation}
under the constraint of a Margin Ranking Loss:
\begin{equation} 
\mathcal{L}_{\text{Rank}} = \frac{1}{|\mathcal{P}k|} \sum_{(j, i) \in \mathcal{P}_k} \max\left(0, \tau_i - \tau_j + m\right),
\label{eq:lrank}
\end{equation}
where $m$ is the margin, and $\mathcal{P}_k = \{(j, i) \mid j > i, \ j, i \in \{1,2,\dots,N\}\}$.
This relative time serves as a critical condition for the generative process, signaling the phase of the contrast enhancement.

We model the distribution of continuous latent representations using the designed HCVAE. The core idea is that dynamic enhancement is a function of both tissue type and time. Therefore, the generation of a specific dynamic feature $\hat{X}_{d}(\vect{p})$ at any position $\vect{p}$ is conditioned on two variables: the neighboring anatomic representation $\hat{X}_{s}(\vect{p})$, which provides the structural context, and the relative time $\tau_i$, which provides the phase context.

The HCVAE's objective is to learn a probabilistic mapping for the dynamic features, which consists of an encoder and a decoder. The encoder $q_{\psi}$ approximates the posterior Gaussian distribution $p(z(\vect{p}) | X_d(\vect{p}), \hat{X}_s(\vect{p}), \tau)$:
\begin{equation}
\begin{split}
z(\vect{p}) &\sim q_{\psi}(z(\vect{p}) | X_d(\vect{p}), \hat{X}_s(\vect{p}), \tau) \\
& = \mathcal{N}(z(\vect{p}); \mu_{\psi}(\vect{p}), \sigma^2_{\psi}(\vect{p})I),
\end{split}
\end{equation}
by predicting its mean and variance:
\begin{equation}
\begin{split}
\mu_{\psi}(\vect{p}) &= \mu_{\psi}(X_d(\vect{p}), \hat{X}_s(\vect{p}), \tau) \\
\sigma^2_{\psi}(\vect{p}) &= \sigma^2_{\psi}(X_d(\vect{p}), \hat{X}_s(\vect{p}), \tau).
\end{split}
\end{equation}
The decoder $p_{\theta}$ then reconstructs the original dynamic feature for each modality $\hat{X}_{d,i}(\vect{p})$ from the sampled and reparameterized latent variable $z$ and the respective conditions $\vect{c}_i(\vect{p}) = (\hat{X}_s(\vect{p}), \tau_i)$:
\begin{equation}
\hat{X}_{d,i}(\vect{p}) = p_{\theta}(X_{d,i}(\vect{p}) | z_{i}(\vect{p}), \vect{c}_i(\vect{p})).
\label{eq:cvae_dec}
\end{equation}

During training, this HCVAE module is trained with KL Divergence and reconstruction loss. The total loss of the modal-specific path is:
\begin{align}
\begin{split}
\mathcal{L}_{\text{Spe}} = & \mathcal{L}_{\text{Rank}} + \underbrace{\mathbb{E}_{q_{\psi}}[-\log p_{\theta}(X_d | z, \vect{c})]}_{\text{Reconstruction Loss}} \\
& + \lambda \underbrace{D_{KL}(q_{\psi}(z | X_d, \vect{c}) \| p_{\theta}(z | \vect{c}))}_{\text{KL Divergence}}
\end{split}
\label{eq:loss_cvae}
\end{align}
where $p_{\theta}(z | \vect{c})$ is the conditional prior, which is a centered isotropic multivariate Gaussian $p_\theta(z|\vect{c})=\mathcal{N}(z;0, I)$.

Crucially, the HCVAE is conditioned on the output of the modal-agnostic module and the known phase. This aligns with our physical model where the dynamic enhancement $H_d$ depends on the static tissue $H_s$ and time $\tau$ in Eq.~(\ref{eq:hypo}). By training this HCVAE, it is capable of capturing the dynamic enhancement properties at position $\vect{p}$ for a specific phase $\tau$. This generative design provides a powerful mechanism for handling missing modalities. For flexible input modalities in the inference stage, a relative time $\tau_i$ can be predicted for each modality. And then a corresponding dynamic representation $\hat{X}_{d,i}$ can be generated by sampling equivalent $z_i$ from the prior distribution and decoding it with the inferred anatomy $\hat{X}_{s}$ and the target time $\tau_i$.

\subsection{Adaptive Assembly}
\label{sec:assembly}

The disentangled representations $\hat{X}_s \in \reals^{C \times K}$ and $\hat{X}_{d} \in \reals^{N \times C \times K}$ capture complementary information. Because of the uncertainty of $N$, the representations cannot be simply concatenated and sent into the fixed-channel decoder. Moreover, their relative importance for the final segmentation task might vary. 

We therefore propose an adaptive assembling module that learns to optimally weight and combine the representations. This module is implemented with an attention mechanism, which computes the attention weights at each position $\vect{p}$. For the static and dynamic representations for all modalities $\hat{X}_m=\{\hat{X}_s, \hat{X}_{d}\}, m \in \{1,2,\dots,N+1\}$, we calculate the attention score with:
\begin{equation}
\alpha_m(\vect{p}) = \underset{m}{\operatorname{Softmax}}(\operatorname{FFN}(\hat{X}_m(\vect{p})))
\label{eq:assemble_gate}
\end{equation}
where FFN means a shared small feed-forward network. This FFN regresses an importance score for each representation at $\vect{p}$. And then the importance scores are normalized as the attention weights. It's worth noting that the attention weights can vary at different positions, which enables the model to decide which modality to focus on when classifying a certain voxel. The final fused representation $\hat{X}(\vect{p})$ is a weighted sum:
\begin{equation}
X_N(\vect{p}) = \sum_{m=1}^{N+1} \alpha_m(\vect{p}) \hat{X}_m(\vect{p}).
\label{eq:assemble_fuse}
\end{equation}
This $X_N \in \reals^{1 \times C \times K}$ is then reshaped as $F_N$ and passed to a segmentation decoder to produce the final mask.

\subsection{Optimization Objectives}
\label{sec:loss}

The entire model is trained end-to-end by optimizing a composite loss function $\Ltotal$, which is a weighted sum of four components:
\begin{equation}
\Ltotal = \mathcal{L}_{\text{Agn}} + \mathcal{L}_{\text{Spe}} + \mathcal{L}_{\text{DE}} + \Lseg,
\label{eq:loss_total}
\end{equation}
where $\mathcal{L}_{\text{DE}}$ is the disentanglement loss and $\Lseg$ is the segmentation loss. Following \cite{zhang2024prototypical}, we use CLUB loss \cite{cheng2020club} to disentangle the representations by mutual information (MI) minimization, which provides a contrastive upper bound on MI, preventing information leakage between the anatomical and temporal latent spaces. Specifically, we calculate
\begin{equation}
    \mathcal{L}_{\text{DE}} = \sum_{i \neq j} \mathcal{L}_{\text{CLUB}}(\hat{X}_{m, i}, \hat{X}_{m, j})
\label{eq:ldisen}
\end{equation}
to minimize the MI not only between the static and dynamic representations, but also among multi-modal dynamic representations. $\Lseg$ is the primary task loss, driving the final segmentation. We use a standard combination of Dice and Cross-Entropy (CE) loss:
\begin{equation}
\Lseg = \mathcal{L}_{\text{Dice}}(S, \hat{S}) + \mathcal{L}_{\text{CE}}(S, \hat{S})
\label{eq:loss_seg}
\end{equation}
where $S$ and $\hat{S}$ are the ground-truth and predicted segmentation masks, respectively.

\subsection{Handling Variable Modality Sequences}
\label{sec:strategy}
A core strength of TARDis is its ability to process an arbitrary number of input modalities $N$, where $N$ varies across different patients and clinical scenarios. This variability presents challenges for standard batch-based optimization. We address this through a sequence-based training strategy and specific inference protocols. The variance of $N$ is taken into consideration in every part of training and inference, including data loading, representation encoding, multi-modal assembly (Sec.~\ref{sec:assembly}), and loss calculation.

\subsubsection*{Sequence-Based Optimization}
In standard deep learning frameworks, input tensors typically require fixed dimensions across a batch. Since our input is a set of volumes $I_{N}$ where $N$ is variable, we treat the input batch as a collection of independent sequences rather than a rigid tensor. Let a training batch $\mathcal{B}$ consist of $B$ subjects, where the $b$-th subject possesses $N_b$ modalities. The total loss is computed by aggregating sample-wise losses, normalized by the sequence length of each subject:
\begin{equation}
    \mathcal{L}_{batch} = \frac{1}{B} \sum_{b=1}^{B} \left( \frac{1}{\omega(N_b)} \mathcal{L}_{total}^{(b)} \right),
\end{equation}
where $\omega(N_b)$ is a normalization factor accounting for the combinatorial number of pairs in the consistency, ranking, and CLUB losses (Eq.~\ref{eq:lconsis},\ref{eq:lrank},\ref{eq:ldisen}). This ensures that subjects with more available modalities do not dominate the gradient updates. %We enforce a training constraint of $N_b \ge 2$ to enable the computation of these pairwise disentanglement objectives.

\subsubsection*{Modality Dropout}
To simulate the missing modality problem explicitly during training, we employ an aggressive \textit{Modality Dropout} strategy. Given a subject with a complete set of multi-phase acquisitions, we randomly subset the inputs to a smaller size $N' \le N$ before passing them to the shared encoder. This stochastic dropping prevents the model from relying on any specific combination of phases and forces the shared encoder to extract robust anatomical features from sparse temporal samples.

\subsubsection*{Generative Inference}
During inference, the model adapts to the specific subset of available modalities without architectural modification. For the modal-agnostic path, available inputs are encoded and queried against the dictionary to form $\hat{X}_s$. For the modal-specific path, we leverage the generative nature of the CVAE. The latent variables $z(p)$ are sampled directly from the prior distribution $\mathcal{N}(0, I)$. These are then decoded, conditioned on the inferred anatomy $\hat{X}_s$ and the estimated relative time $\tau$, to reconstruct the probabilistic dynamic representations $\hat{X}_d$. This allows TARDis to understand the grounding of the tumor even in cases of extreme data sparsity (e.g., $N=1$).

\section{Results}
\label{sec:results}

% \subsection{Experimental Setting}
% modality distribution
\begin{figure}[tb]
  \centering
  % \fbox{\rule{0pt}{3in} \rule{0.95\columnwidth}{0pt}}
  \includegraphics[width=0.5\columnwidth]{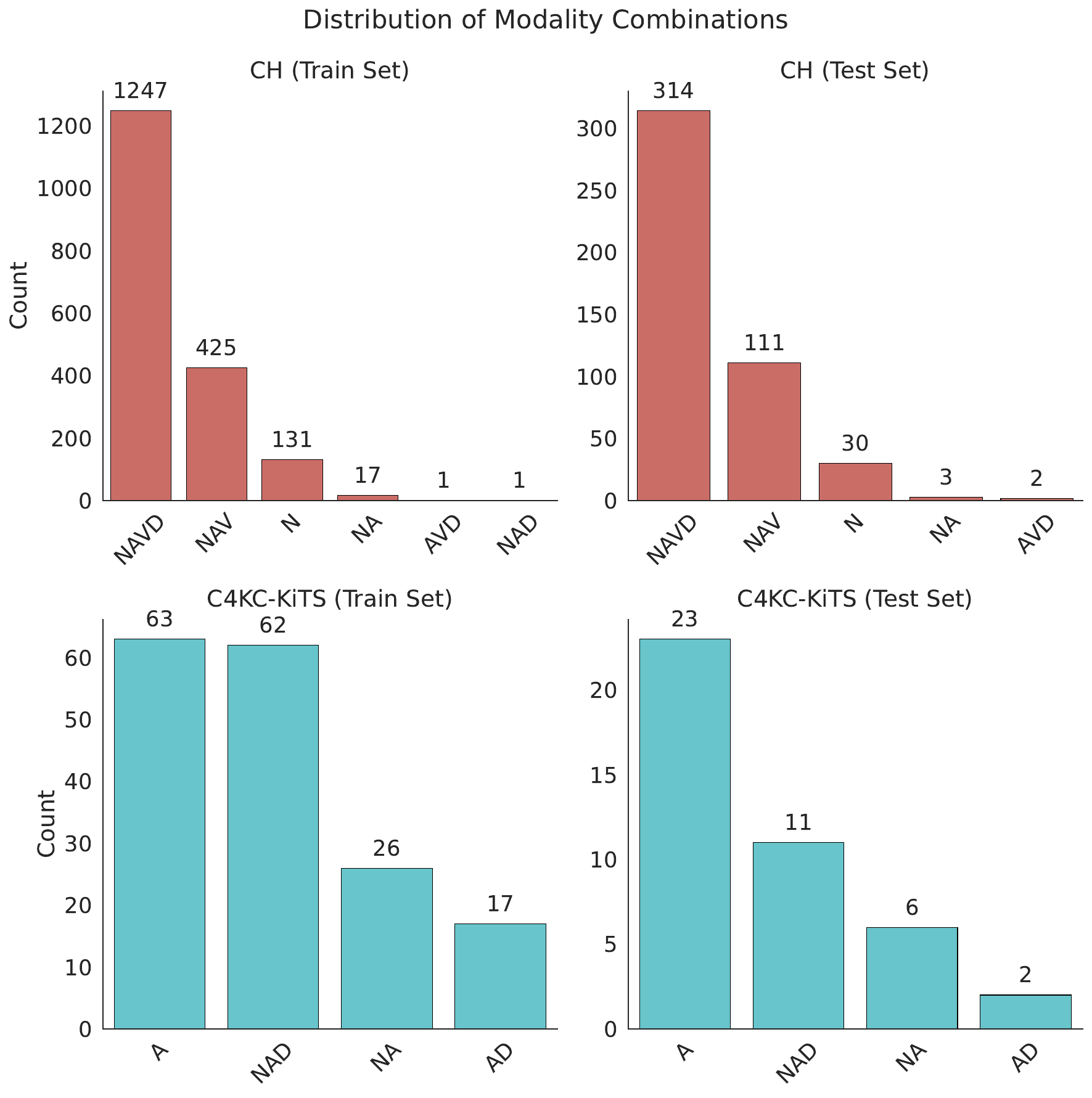}
  \caption{The distribution of modality combinations in CT datasets. N: Non-Contrast Scan. A: Arterial Phase. V: Portal Venous Phase. D: Delayed Phase. CH: ChangHai Dataset.}
  \label{fig:distribution}
\end{figure}

\begin{figure*}[tb]
  \centering
  % \fbox{\rule{0pt}{3in} \rule{0.95\columnwidth}{0pt}}
  \includegraphics[width=\textwidth]{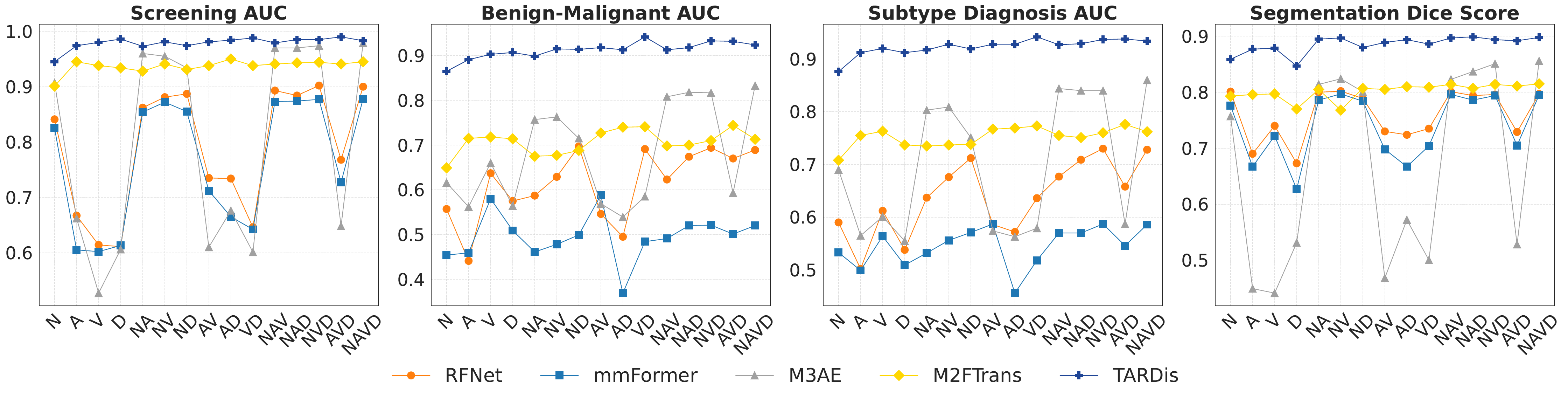}
  \caption{The metrics curves of compared methods with different modality combinations in CH dataset. N: Non-Contrast Scan. A: Arterial Phase. V: Portal Venous Phase. D: Delayed Phase.}
  \label{fig:metrics_CH}
\end{figure*}

\begin{table*}[tb]
    \centering
    \caption{Classification performance for tumor screening, benign-malignant classification, and diagnosis in CH dataset}
    \label{tab:auc_CH}
    \resizebox{\textwidth}{!}{%
    \begin{tabular}{cccc|ccccc|ccccc|ccccc}
        \toprule
        \multicolumn{4}{c}{\textbf{Modality}}  & \multicolumn{5}{c}{\textbf{Screening AUC}} & \multicolumn{5}{c}{\textbf{Benign-Malignant AUC}} & \multicolumn{5}{c}{\textbf{Subtype Diagnosis AUC}} \\
        \cmidrule(r){1-4} \cmidrule(lr){5-9} \cmidrule(lr){10-14} \cmidrule(l){15-19}
        \textbf{N} & \textbf{A} & \textbf{V} & \textbf{D} & RFNet & mmFormer & M3AE & M2FTrans & TARDis & RFNet & mmFormer & M3AE & M2FTrans & TARDis & RFNet & mmFormer & M3AE & M2FTrans & TARDis \\
        \midrule
        $\surd$ & & & & 0.841  & 0.825  & 0.907  & 0.901  & \textbf{0.945} & 0.557  & 0.454  & 0.616  & 0.649  & \textbf{0.865}  & 0.590  & 0.533  & 0.690  & 0.708 & \textbf{0.876}  \\
        & $\surd$ & & & 0.667  & 0.605  & 0.662  & 0.945  & \textbf{0.974}  & 0.441  & 0.459  & 0.562  & 0.715  & \textbf{0.891}  & 0.502  & 0.499  & 0.565  & 0.755  & \textbf{0.912} \\
        & & $\surd$ & & 0.614  & 0.602  & 0.527  & 0.938  & \textbf{0.980}  & 0.637  & 0.580  & 0.660  & 0.718  & \textbf{0.903}  & 0.612  & 0.564  & 0.601  & 0.763  & \textbf{0.920} \\
        & & & $\surd$ & 0.611  & 0.613  & 0.606  & 0.934  & \textbf{0.986}  & 0.575  & 0.509  & 0.564  & 0.714  & \textbf{0.907}  & 0.538  & 0.509  & 0.555  & 0.737  & \textbf{0.912}  \\
        $\surd$ & $\surd$ & & & 0.862  & 0.854  & 0.960  & 0.928  & \textbf{0.973}  & 0.587  & 0.461  & 0.757  & 0.675  & \textbf{0.899}  & 0.637  & 0.532  & 0.803  & 0.735  & \textbf{0.917}  \\
        $\surd$ & & $\surd$ & & 0.881  & 0.872  & 0.955  & 0.941  & \textbf{0.981}  & 0.629  & 0.478  & 0.763  & 0.677  & \textbf{0.915}  & 0.676  & 0.556  & 0.809  & 0.737  & \textbf{0.928} \\
        $\surd$ & & & $\surd$ & 0.887  & 0.855  & 0.934  & 0.931  & \textbf{0.974}  & 0.697  & 0.499  & 0.715  & 0.688  & \textbf{0.914}  & 0.712  & 0.571  & 0.751  & 0.738  & \textbf{0.919}  \\
        & $\surd$ & $\surd$ & & 0.735  & 0.712  & 0.610  & 0.938  & \textbf{0.981}  & 0.546  & 0.588  & 0.569  & 0.727  & \textbf{0.918}  & 0.586  & 0.587  & 0.574  & 0.767  & \textbf{0.928} \\
        & $\surd$ & & $\surd$ & 0.734  & 0.665  & 0.676  & 0.950  & \textbf{0.984}  & 0.495  & 0.369  & 0.539  & 0.740  & \textbf{0.913}  & 0.572  & 0.456  & 0.563  & 0.769  & \textbf{0.928}  \\
        & & $\surd$ & $\surd$ & 0.646  & 0.642  & 0.601  & 0.938  & \textbf{0.988}  & 0.691  & 0.484  & 0.585  & 0.741  & \textbf{0.942}  & 0.636  & 0.518  & 0.579  & 0.773  & \textbf{0.942}  \\
        $\surd$ & $\surd$ & $\surd$ & & 0.893  & 0.873  & 0.970  & 0.941  & \textbf{0.979}  & 0.623  & 0.491  & 0.808  & 0.698  & \textbf{0.913}  & 0.677  & 0.570  & 0.844  & 0.755  & \textbf{0.927} \\
        $\surd$ & $\surd$ & & $\surd$ & 0.884  & 0.874  & 0.970  & 0.943  & \textbf{0.985}  & 0.674  & 0.520  & 0.818  & 0.700  & \textbf{0.918}  & 0.709  & 0.570  & 0.840  & 0.751  & \textbf{0.929}  \\
        $\surd$ & & $\surd$ & $\surd$ & 0.902  & 0.877  & 0.974  & 0.944  & \textbf{0.985}  & 0.694  & 0.521  & 0.817  & 0.710  & \textbf{0.933}  & 0.730  & 0.587  & 0.840  & 0.760  & \textbf{0.937} \\
        & $\surd$ & $\surd$ & $\surd$ & 0.768  & 0.727  & 0.648  & 0.941  & \textbf{0.990}  & 0.670  & 0.501  & 0.593  & 0.744  & \textbf{0.932}  & 0.658  & 0.546  & 0.587  & 0.776  & \textbf{0.938} \\
        $\surd$ & $\surd$ & $\surd$ & $\surd$ & 0.900  & 0.878  & 0.979  & 0.945  & \textbf{0.983}  & 0.689  & 0.520  & 0.833  & 0.713  & \textbf{0.924}  & 0.728  & 0.586  & 0.860  & 0.762  & \textbf{0.934} \\
        \midrule
        \multicolumn{4}{c|}{\textbf{Average}} & 0.788  & 0.765  & 0.799  & 0.937  & \textbf{0.979}  & 0.614  & 0.496  & 0.680  & 0.707  & \textbf{0.912}  & 0.638  & 0.546  & 0.697  & 0.752  & \textbf{0.923} \\
        \bottomrule
    \end{tabular}
    }
\end{table*}

\begin{table}[tb]
    \centering
    \caption{Tumor segmentation performance in CH dataset}
    \label{tab:dice_CH}
    % \resizebox{\columnwidth}{!}{%
    \begin{tabular}{cccc|ccccc}
        \toprule
        \multicolumn{4}{c}{\textbf{Modality}} & \multicolumn{5}{c}{\textbf{Segmentation Dice}} \\
        \cmidrule(r){1-4} \cmidrule(l){5-9}
        \textbf{N} & \textbf{A} & \textbf{V} & \textbf{D} & RFNet & mmFormer & M3AE & M2FTrans & TARDis \\
        \midrule
        $\surd$ & & & & 0.801  & 0.776  & 0.757  & 0.793  & \textbf{0.859} \\
        & $\surd$ & & & 0.690  & 0.667  & 0.449  & 0.796  & \textbf{0.877}  \\
        & & $\surd$ & & 0.740  & 0.722  & 0.441  & 0.797  & \textbf{0.879} \\
        & & & $\surd$ & 0.673  & 0.627  & 0.531  & 0.770  & \textbf{0.847} \\
        $\surd$ & $\surd$ & & & 0.800  & 0.786  & 0.814  & 0.805  & \textbf{0.895} \\
        $\surd$ & & $\surd$ & & 0.802  & 0.797  & 0.824  & 0.768  & \textbf{0.897} \\
        $\surd$ & & & $\surd$ & 0.789  & 0.784  & 0.801  & 0.807  & \textbf{0.880} \\
        & $\surd$ & $\surd$ & & 0.730  & 0.698  & 0.468  & 0.805  & \textbf{0.889} \\
        & $\surd$ & & $\surd$ & 0.724  & 0.667  & 0.572  & 0.810  & \textbf{0.894} \\
        & & $\surd$ & $\surd$ & 0.735  & 0.704  & 0.500  & 0.809  & \textbf{0.886} \\
        $\surd$ & $\surd$ & $\surd$ & & 0.801  & 0.796  & 0.823  & 0.814  & \textbf{0.897} \\
        $\surd$ & $\surd$ & & $\surd$ & 0.794  & 0.786  & 0.837  & 0.807  & \textbf{0.899} \\
        $\surd$ & & $\surd$ & $\surd$ & 0.796  & 0.794  & 0.851  & 0.814  & \textbf{0.894} \\
        & $\surd$ & $\surd$ & $\surd$ & 0.729  & 0.705  & 0.528  & 0.811  & \textbf{0.892} \\
        $\surd$ & $\surd$ & $\surd$ & $\surd$ & 0.797  & 0.795  & 0.856  & 0.815  & \textbf{0.898} \\
        \midrule
        \multicolumn{4}{c|}{\textbf{Average}} & 0.760  & 0.740  & 0.670  & 0.801  & \textbf{0.885} \\
        \bottomrule
    \end{tabular}
    % }
\end{table}

\begin{table}[tb]
    \centering
    \caption{Segmentation and screening performance in C4KC-KiTS dataset}
    \label{tab:kits19}
    \begin{tabular}{c|ccc|ccc}
        \toprule
        \multirow{2}{*}{\textbf{Method}} & \multicolumn{3}{c|}{\textbf{Segmentation}} & \multicolumn{3}{c}{\textbf{Screening}} \\
        \cmidrule(lr){2-4} \cmidrule(l){5-7}
        & \textbf{Dice} & \textbf{Precision} & \textbf{Recall} & \textbf{ACC} & \textbf{SEN} & \textbf{AUC} \\
        \midrule
        RFNet & 0.767 & 0.811 & 0.835 & 0.904 & 0.909 & 0.933 \\
        mmFormer & 0.703 & 0.787 & 0.774 & 0.855 & 0.886 & 0.902 \\
        M3AE & 0.790 & 0.840 & 0.836 & 0.940 & \textbf{0.955} & 0.948 \\
        M2FTrans & 0.321 & 0.394 & 0.675 & 0.735 & 0.682 & 0.773 \\
        TARDis & \textbf{0.825} & \textbf{0.860} & \textbf{0.841} & \textbf{0.952} & \textbf{0.955} & \textbf{0.965} \\
        \bottomrule
    \end{tabular}
\end{table}

\begin{table*}[tb]
    \centering
    \caption{Tumor segmentation performance in BraTS18 dataset}
    \label{tab:brats18}
    \resizebox{\textwidth}{!}{%
    \begin{tabular}{cccc|ccccc|ccccc|ccccc}
        \toprule
        \multicolumn{4}{c}{\textbf{Modality}}  & \multicolumn{5}{c}{\textbf{Whole Tumor Dice}} & \multicolumn{5}{c}{\textbf{Tumor Core Dice}} & \multicolumn{5}{c}{\textbf{Enhancing Tumor Dice}} \\
        \cmidrule(r){1-4} \cmidrule(lr){5-9} \cmidrule(lr){10-14} \cmidrule(l){15-19}
        \textbf{T1} & \textbf{T1CE} & \textbf{T2} & \textbf{FLAIR} & RFNet & mmFormer & M3AE & M2FTrans & TARDis & RFNet & mmFormer & M3AE & M2FTrans & TARDis & RFNet & mmFormer & M3AE & M2FTrans & TARDis \\
        \midrule
        $\surd$ & & & & 0.744  & 0.731  & 0.045  & 0.721  & \textbf{0.746}  & \textbf{0.636}  & 0.599  & 0.056  & 0.593  & \textbf{0.636}  & \textbf{0.330}  & 0.289  & 0.043  & 0.287  & 0.269  \\
        & $\surd$ & & & 0.727  & 0.697  & 0.718  & 0.688  & \textbf{0.747}  & 0.788  & 0.746  & 0.787  & 0.735  & \textbf{0.802}  & 0.723  & 0.616  & 0.668  & 0.650  & \textbf{0.728} \\
        & & $\surd$ & & 0.838  & 0.828  & 0.223  & 0.827  & \textbf{0.849}  & 0.623  & 0.549  & 0.341  & 0.564  & \textbf{0.649}  & 0.367  & 0.293  & 0.230  & 0.317  & \textbf{0.385} \\
        & & & $\surd$ & 0.869  & 0.863  & 0.136  & 0.856  & \textbf{0.879}  & 0.624  & 0.571  & 0.175  & 0.569  & \textbf{0.647}  & 0.304  & 0.260  & 0.076  & 0.271  & \textbf{0.360}  \\
        $\surd$ & $\surd$ & & & 0.754  & 0.755  & 0.624  & 0.744  & \textbf{0.779}  & 0.771  & 0.753  & 0.531  & 0.757  & \textbf{0.820}  & 0.623  & 0.628  & 0.241  & 0.646  & \textbf{0.716}  \\
        $\surd$ & & $\surd$ & & 0.831  & 0.806  & 0.451  & 0.811  & \textbf{0.868}  & 0.675  & 0.615  & 0.323  & 0.577  & \textbf{0.705}  & 0.374  & 0.322  & 0.111  & 0.329  & \textbf{0.393} \\
        $\surd$ & & & $\surd$ & 0.881  & 0.853  & 0.082  & 0.867  & \textbf{0.893}  & 0.683  & 0.636  & 0.143  & 0.651  & \textbf{0.715}  & 0.341  & 0.287  & 0.083  & 0.324  & \textbf{0.359}  \\
        & $\surd$ & $\surd$ & & 0.822  & 0.820  & 0.636  & 0.807  & \textbf{0.866}  & 0.776  & 0.733  & 0.653  & 0.699  & \textbf{0.829}  & 0.633  & 0.614  & 0.570  & 0.609  & \textbf{0.749} \\
        & $\surd$ & & $\surd$ & 0.870  & 0.858  & 0.498  & 0.843  & \textbf{0.899}  & 0.747  & 0.732  & 0.382  & 0.684  & \textbf{0.828}  & 0.575  & 0.598  & 0.147  & 0.510  & \textbf{0.744}   \\
        & & $\surd$ & $\surd$ & 0.888  & 0.877  & 0.081  & 0.876  & \textbf{0.898}  & 0.660  & 0.610  & 0.071  & 0.616  & \textbf{0.682}  & 0.347  & 0.307  & 0.010  & 0.300  & \textbf{0.400}  \\
        $\surd$ & $\surd$ & $\surd$ & & 0.819  & 0.803  & 0.868  & 0.811  & \textbf{0.866}  & 0.749  & 0.727  & 0.652  & 0.725  & \textbf{0.833}  & 0.505  & 0.561  & 0.309  & 0.594  & \textbf{0.749} \\
        $\surd$ & $\surd$ & & $\surd$ & 0.861  & 0.831  & 0.411  & 0.844  & \textbf{0.898}  & 0.752  & 0.730  & 0.481  & 0.736  & \textbf{0.839}  & 0.541  & 0.564  & 0.437  & 0.558  & \textbf{0.719}  \\
        $\surd$ & & $\surd$ & $\surd$ & 0.887  & 0.862  & 0.041  & 0.879  & \textbf{0.902}  & 0.692  & 0.636  & 0.053  & 0.629  & \textbf{0.726}  & 0.368  & 0.311  & 0.050  & 0.286  & \textbf{0.405}  \\
        & $\surd$ & $\surd$ & $\surd$ & 0.882  & 0.872  & 0.742  & 0.862  & \textbf{0.907}  & 0.738  & 0.708  & 0.640  & 0.664  & \textbf{0.836}  & 0.490  & 0.540  & 0.332  & 0.439  & \textbf{0.748} \\
        $\surd$ & $\surd$ & $\surd$ & $\surd$ & 0.880  & 0.845  & 0.844  & 0.860  & \textbf{0.906}  & 0.751  & 0.705  & 0.664  & 0.696  & \textbf{0.841}  & 0.476  & 0.525  & 0.359  & 0.473  & \textbf{0.740} \\
        \midrule
        \multicolumn{4}{c|}{\textbf{Average}} & 0.837  & 0.820  & 0.427  & 0.820  & \textbf{0.860}  & 0.711  & 0.670  & 0.397  & 0.660  &\textbf{ 0.759}  & 0.466  & 0.448  & 0.244  & 0.439  & \textbf{0.564} \\
        \bottomrule
    \end{tabular}
    }
\end{table*}

\begin{figure*}[tb]
  \centering
  % \fbox{\rule{0pt}{3in} \rule{0.95\columnwidth}{0pt}}
  \includegraphics[width=\textwidth]{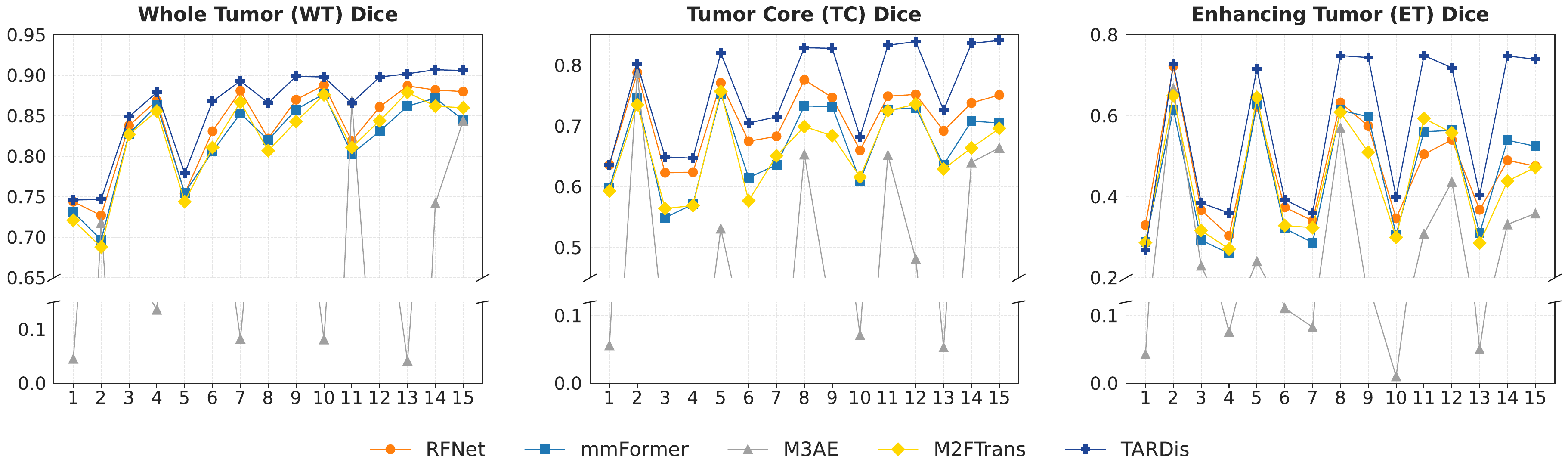}
  \caption{The metrics curves of compared methods with different modality combinations in the BraTS18 dataset. The number on the x-axis refers to the modality combination in Table \ref{tab:brats18}.}
  \label{fig:metrics_brats}
\end{figure*}

\begin{figure}[tb]
  \centering
  % \fbox{\rule{0pt}{3in} \rule{0.95\columnwidth}{0pt}}
  \includegraphics[width=0.9\columnwidth]{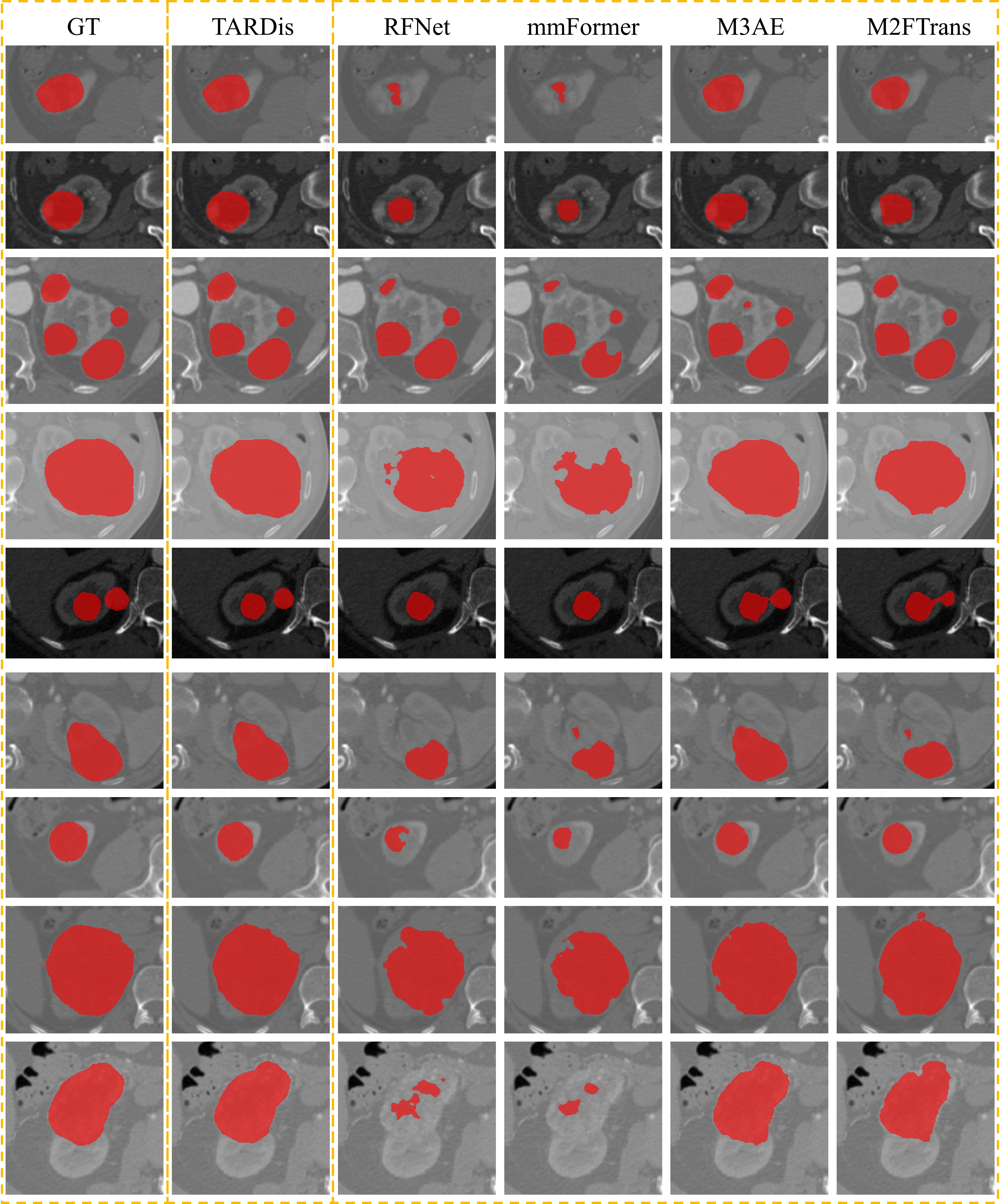}
  \caption{Qualitative comparison results for kidney tumor segmentation.}
  \label{fig:vis_compare}
\end{figure}

\begin{figure}[tb]
  \centering
  % \fbox{\rule{0pt}{3in} \rule{0.95\columnwidth}{0pt}}
  \includegraphics[width=0.9\columnwidth]{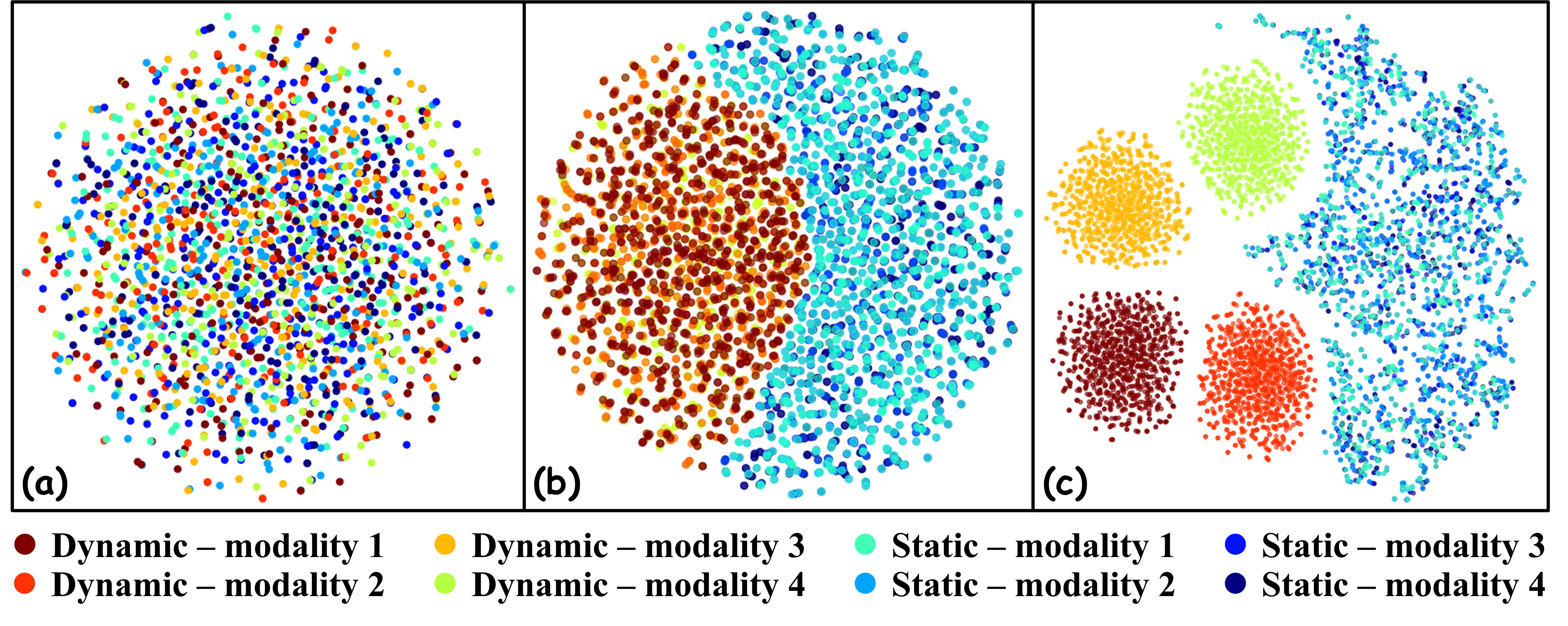}
  \caption{The t-SNE plots of features. (a) Entangled features are inseparable. (b) If we only use two branches to reconstruct the static and dynamic features respectively, the features are separated into two distinct clusters (left - dynamic/right - static). (c) If we further use $\mathcal{L}_{\text{Rank}}$ and $\mathcal{L}_{\text{DE}}$ as constraints, the multi-modal dynamic features will also be separated into sub-clusters.}
  \label{fig:tsne}
\end{figure}
\subsection{Datasets}
To evaluate the efficacy and robustness of TARDis, we utilized three datasets comprising both Computed Tomography (CT) and Magnetic Resonance Imaging (MRI) modalities. The distribution of modality combinations for the CT datasets is illustrated in Fig.~\ref{fig:distribution}.

\subsubsection*{Changhai (CH) Dataset}
To validate our method in a large-scale multi-modal clinical setting, we collected a private multi-phase abdominal CT dataset from Changhai Hospital, Shanghai, China. This extensive cohort contains 2,282 patients comprising 8,065 total scans (Fig.~\ref{fig:distribution}). The dataset includes 2,958 annotated kidney lesions spanning 26 distinct histological subtypes. For experimental purposes, these subtypes were categorized into four clinically relevant classes based on surgical management strategies: cysts (2,188), other benign lesions (105), low-grade malignant lesions (450), and high-grade malignant lesions (215). All volumes were spatially registered and preprocessed to facilitate multi-modal tumor diagnosis. Ground truth labels were annotated and iteratively refined by board-certified radiologists.

\subsubsection*{C4KC-KiTS}
The first dataset is the \textit{Climb 4 Kidney Cancer - Kidney and Kidney Tumor Segmentation} (C4KC-KiTS) collection~\cite{C4KCKITS2019,heller2019kits19,heller2020state}, a public subset of the KiTS19 challenge data. We utilized 210 patients from the training set. As shown in Fig.~\ref{fig:distribution}, every patient includes an arterial phase scan, while a subset of patients also contains non-contrast or delayed phase acquisitions, providing a controlled environment for testing missing-modality performance.

\subsubsection*{BraTS 2018}
While TARDis is theoretically grounded in the CT modalities, we extended our evaluation to the BraTS 2018 dataset~\cite{menze2014multimodal, bakas2017advancing,bakas2018identifying} to compare with existing state-of-the-art (SOTA) incomplete modality frameworks, which are predominantly established on this neuroimaging standard. BraTS18 provides four co-registered MRI modalities per patient: T1-weighted (T1), post-contrast T1-weighted (T1CE), T2-weighted (T2), and FLAIR. Annotations include three nested sub-regions: GD-enhancing tumor (ET), peritumoral edema (ED), and the necrotic/non-enhancing tumor core (NCR/NET). Following standard challenge protocols, we evaluated performance on three aggregated regions: Whole Tumor (WT), Tumor Core (TC), and Enhancing Tumor (ET).

\subsection{Tasks and Metrics}
We designed a comprehensive evaluation protocol covering segmentation, screening, and diagnosis.

\subsubsection*{CH Dataset} We implemented a hierarchical pipeline consisting of three tasks: (1) \textbf{Tumor Screening}, binary detection of lesion presence; (2) \textbf{Benign-Malignant Classification}; and (3) \textbf{Subtype Diagnosis}, classification into the four categories described above. Performance for these classification tasks was quantified using the Area Under the Receiver Operating Characteristic Curve (AUC). Additionally, \textbf{Tumor Segmentation} performance was evaluated using the Dice Similarity Coefficient.

\subsubsection*{C4KC-KiTS} We evaluated \textbf{Tumor Segmentation} using Dice, Precision, and Recall. We also assessed \textbf{Screening} performance using Accuracy (ACC), Sensitivity (SEN), and AUC.

\subsubsection*{BraTS 2018} Consistent with the literature, performance was evaluated solely on \textbf{Tumor Segmentation} (WT, TC, ET) using the Dice Similarity Coefficient.

\subsection{Implementation Details}
% rand drop
We compared TARDis against four leading methods in incomplete multi-modal learning: RFNet~\cite{ding2021rfnet}, mmFormer~\cite{zhang2022mmformer}, M3AE~\cite{liu2023m3ae}, and M2FTrans~\cite{m2ftrans}.

All experiments were conducted using PyTorch 1.12 on NVIDIA H20 GPUs. We utilized the AdamW optimizer with a decreasing learning rate schedule initialized at 0.01. The batch size was set to 4, employing the sequence-based optimization strategy detailed in Sec.~\ref{sec:strategy}. To simulate missing modalities during training, we applied a stochastic modality dropout with a rate of 20\%, while ensuring that at least one modality remained available for every training sample. The embedding dictionary size was fixed at 512. The weight $\beta$ in Eq.~\ref{eq:lagn} was set to 0.25, following \cite{van2017neural}. When training with the BraTS18, the ranking loss $\mathcal{L}_{\text{Rank}}$ was replaced with a separation loss that attempts to push the $\tau$ far from each other. All baseline models were retrained and evaluated under identical conditions to ensure fair comparison.

\subsection{Comparisons with the State-of-the-Art Methods}
% CH, kits19

We compared TARDis against four state-of-the-art frameworks. The quantitative results across the CH, C4KC-KiTS, and BraTS18 datasets are summarized in Tables \ref{tab:auc_CH}, \ref{tab:dice_CH}, \ref{tab:kits19}, and \ref{tab:brats18}. On the large-scale multi-modal CH dataset, TARDis consistently outperformed all baselines across both classification and segmentation tasks. As shown in Table \ref{tab:auc_CH}, our method achieved an average Screening AUC of 0.979, significantly surpassing the second-best method, M2FTrans (0.937). Notably, in the challenging Subtype Diagnosis task, TARDis demonstrated remarkable superiority with an AUC of 0.923, whereas competing methods struggled to exceed 0.760. This substantial margin suggests that explicitly disentangling the dynamic tumor enhancement significantly aids in distinguishing subtle histological subtypes.

For segmentation tasks, Table \ref{tab:dice_CH} reveals that TARDis maintains high accuracy even under severe data sparsity. While baseline performance degrades sharply when the arterial or venous phases are missing, TARDis maintains a robust Dice score of 0.859 on the non-contrast scan alone. This indicates our model's capability to infer missing hemodynamic information from the learned static anatomy. Similarly, on the public C4KC-KiTS dataset (Table \ref{tab:kits19}), TARDis achieved the highest scores across all metrics, with a Segmentation Dice of 0.825 and a Screening AUC of 0.965.

Although theoretically grounded in CT hemodynamics, TARDis generalizes effectively to MRI modalities. As evidenced in Table \ref{tab:brats18}, our framework achieves the highest average Dice scores for Whole Tumor (0.860), Tumor Core (0.759), and the challenging Enhancing Tumor (0.564). Specifically, for the Enhancing Tumor region, which relies heavily on contrast-enhanced sequences, TARDis outperforms M3AE and RFNet by margins of roughly 12\% and 9\%, respectively. This confirms that the proposed disentanglement effectively captures temporal-like dependencies between MRI sequences.

All state-of-the-art baselines struggled with the inconvenient design of using separate encoders that require the full set of input modalities to effectively capture the cross-modality information. Their performance severely degrades when facing incomplete inputs. Furthermore, these models, particularly those using transformer-based architectures, often suffer when trained with aggressively large modality dropout rates, which are necessary to simulate clinical sparsity. Our proposed TARDis architecture, however, addresses this by utilizing a Disentangled Time Attenuated Representation that separates static and dynamic features, enabling robust information inference even when key modalities are missing. This fundamental design difference allows TARDis to maintain high performance under challenging, sparse training conditions that cause instability in competing methods.

\subsection{Modality Ablation and Robustness Analysis}
% line gram
To investigate the impact of different modality combinations and verify the robustness of TARDis against missing data, we analyzed the performance variations across all possible subsets of inputs, as visualized in Figs.~\ref{fig:metrics_CH} and \ref{fig:metrics_brats}.

In kidney tumor diagnosis, contrast-enhanced phases are clinically critical. Fig.~\ref{fig:metrics_CH} illustrates that baseline methods exhibit jagged performance curves, with deep troughs occurring whenever the 'A' or 'V' phases are absent. In contrast, TARDis exhibits a much smoother performance curve. For instance, in the Benign-Malignant classification task, standard methods drop to AUCs below 0.60 when limited to non-contrast data. TARDis, however, leverages the CVAE-based modal-specific path to infer the missing perfusion features, maintaining an AUC above 0.85 even in the absence of contrast media.

The missing modality problem is most severe when only a single modality is available. Table \ref{tab:dice_CH} highlights that under the single-modality setting (average of N, A, V, D individual performance), TARDis achieves an average Dice of roughly 0.86, whereas RFNet and mmFormer lag significantly behind at approximately 0.74 and 0.63, respectively. This robustness is further corroborated by Fig.\ref{fig:metrics_brats} for the BraTS dataset, where the variance in Dice scores across different combinations is visibly lower for TARDis compared to baselines. This stability implies that our Disentangled Time Attenuated Representation effectively decouples the diagnostic signal from the modality availability, ensuring reliable clinical deployment regardless of the acquisition protocol.

\subsection{Qualitative Analysis and Visualization}

To intuitively understand the efficacy of TARDis, we present qualitative segmentation results under the most challenging scenario where only a single modality is available (Fig.~\ref{fig:vis_compare}). In these cases, standard missing-modality frameworks such as RFNet, mmFormer, and M3AE struggle significantly. As observed in the visual comparisons, these baseline methods often produce fragmented segmentations or fail to detect the tumor entirely, particularly when the highly informative contrast-enhanced phases (arterial or venous) are absent. This failure likely stems from their reliance on channel fusion or masking strategies; when a specific channel is missing, the corresponding diagnostic signal is effectively lost, and the model cannot compensate for the missing hemodynamic information. In contrast, TARDis maintains high structural consistency with the ground truth across all single-modality inputs. By leveraging the CVAE-based modal-specific path, our model effectively infers the missing perfusion features conditioned on the available static anatomy and the inferred relative time. This capability ensures that the diagnostic boundaries remain distinct even in non-contrast scans, demonstrating the robustness of our generative approach against extreme data sparsity.

We further validate the quality of the learned representations through t-SNE visualization of the latent feature space in Fig.~\ref{fig:tsne}. Fig.~\ref{fig:tsne}(a) illustrates that without specific disentanglement mechanisms, the feature distributions are entangled, making it difficult to distinguish between anatomical and physiological variations. However, by employing our proposed dual-path architecture, a clear separation emerges between the static and dynamic components, as shown in Fig~\ref{fig:tsne}(b). Furthermore, the incorporation of the ranking and disentanglement objectives ($\mathcal{L}_{\text{Rank}}$ and $\mathcal{L}_{\text{DE}}$) refines the dynamic latent space, organizing the features into distinct sub-clusters that correspond to different contrast phases (Fig~\ref{fig:tsne}(c)). This structured latent space confirms that TARDis successfully decouples the time-invariant anatomical information from the time-dependent hemodynamic variations, verifying that the model has learned a physically grounded representation rather than simply memorizing statistical correlations.

\section{Conclusion}
\label{sec:conc}
In this paper, we presented TARDis, a unified framework for incomplete multi-modal tumor segmentation and classification that moves beyond standard channel-masking strategies. By grounding our architectural design in the physical principles of CT hemodynamics, we hypothesized that the diagnostic signal is composed of a static anatomical baseline and a dynamic perfusion process. The integration of a dictionary-based anatomical encoder ensures structural consistency across scans, while the time-conditioned HCVAE allows for the probabilistic reconstruction of dynamic features, even when specific contrast phases are physically absent. Our comprehensive evaluation across large-scale multi-modal CT and MRI datasets confirms that this physics-aware approach yields superior generalization compared to existing methods that treat modalities as uncorrelated inputs. Specifically, the qualitative and quantitative improvements observed in single-modality inference scenarios suggest that TARDis can effectively infer missing hemodynamic information from learned latent priors, which validates our hypothesis. Clinically, by relieving the dependency on complete multi-phase series, TARDis paves the way for flexible acquisition protocols. This capability creates a significant opportunity to reduce scan durations and minimize patient radiation exposure, thereby improving overall hospital efficiency and throughput without compromising diagnostic precision. Furthermore, we believe this continuous-time disentanglement paradigm can be generalized beyond segmentation to broader medical challenges, such as low-dose diagnostic and cross-center data harmonization.

\bibliographystyle{unsrt}
\bibliography{references}

\end{document}